\documentclass[letterpaper, 10 pt, conference,twosiede]{ieeeconf}  

\IEEEoverridecommandlockouts                              

\usepackage{amsmath} 
\usepackage{graphicx}
\usepackage{hyperref}
\usepackage{caption}
\usepackage{subcaption}
\usepackage{multirow}
\usepackage[T1]{fontenc}
\usepackage{array}
\usepackage{float}
\usepackage{textcomp}
\usepackage{amsmath}
\usepackage{amssymb}
\usepackage{stackrel}
\usepackage{graphicx}
\usepackage{lipsum}
\usepackage[utf8]{inputenc}
\usepackage[english]{babel}
\usepackage{fancyhdr}
\begin{document}
\title{\LARGE \bf
Rigid-Soft Interactive Learning for Robust Grasping*
}
\author{Linhan Yang$^{1,\#}$, Fang Wan$^{2,\#}$, Haokun Wang$^{3}$, Xiaobo Liu$^{3}$, Yujia Liu$^{3}$, Jia Pan$^{4}$ and Chaoyang Song$^{5,*}$
\thanks{Manuscript received: September 10, 2019; Revised: December 13, 2019; Accepted: January 9, 2020. This paper was recommended for publication by Editor Kyu-Jin Cho upon evaluation of the Associate Editor and Reviewers’ comments.}
\thanks{*This work was supported by Southern University of Science and Technology and AncoraSpring Inc. $^{\#}$Linhan Yang and Fang Wan are co-first-authors of this paper.}
\thanks{$^{1}$Linhan Yang is a Joint Ph.D. candidate with Southern University of Science and Technology and the University of Hong Kong,
        Shenzhen, Guangdong 518055, China.
        {\tt\small 11950013@mail.sustech.edu.cn}}%
\thanks{$^{2}$Fang Wan is with AncoraSpring, Inc. and SUSTech Institute of Robotics, Southern University of Science and Technology, 
        Shenzhen, Guangdong 518055, China. 
        {\tt\small sophie.fwan@gmail.com}}%
\thanks{$^{3}$Haokun Wang, Xiaobo Liu, and Yujia Liu are with the Department of Mechanical and Energy Engineering, Southern University of Science,
        Shenzhen Guangdong 518055, China. 
        {\tt\small {11510135,1193080,liuyj}@mail.sustech.edu.cn}}%
\thanks{$^{4}$Jia Pan is with Department of Computer Science, University of Hong Kong, 
        Hong Kong 999077. 
        {\tt\small jpan@cs.hku.hk}}
\thanks{$^{5}$Chaoyang Song is the corresponding author with the Department of Mechanical and Energy Engineering, Southern University of Science and Technology,
        Shenzhen, Guangdong 518055, China.
        {\tt\small songcy@ieee.org}}
\thanks{Digital Object Identifier (DOI): see top of this page}
        }
\pagestyle{fancy}
\fancyhf{}
\makeatother
\maketitle
\setlength{\topskip}{5mm}
\begin{abstract}
Inspired by widely used soft fingers on grasping, we propose a method of rigid-soft interactive learning, aiming at reducing the time of data collection. In this paper, we classify the interaction categories into Rigid-Rigid, Rigid-Soft, Soft-Rigid according to the interaction surface between grippers and target objects. We find experimental evidence that the interaction types between grippers and target objects play an essential role in the learning methods. We use soft, stuffed toys for training, instead of everyday objects, to reduce the integration complexity and computational burden and exploit such rigid-soft interaction by changing the gripper fingers to the soft ones when dealing with rigid, daily-life items such as the Yale-CMU-Berkeley (YCB) objects. With a small data collection of 5K picking attempts in total, our results suggest that such Rigid-Soft and Soft-Rigid interactions are transferable. Moreover, the combination of different grasp types shows better performance on the grasping test. We achieve the best grasping performance at 97.5\% for easy YCB objects and 81.3\% for difficult YCB objects while using a precise grasp with a two-soft-finger gripper to collect training data and power grasp with a four-soft-finger gripper to test.
\end{abstract}
\section{Introduction}
\label{sec:Introduction}
Robot learning is widely accepted by academia and industry with its potentials to transform autonomous robot control through machine learning. Recent literature has demonstrated that with the availability of more data, the robot learning method is not only possible but also preferred in some scenarios \cite{bohg2013data}. For example, \cite{pinto2016curious, correll2016analysis, pinto2016supersizing, yahya2017collective} have shown that with the scale-up of data collection using a self-supervised method, grasping success rate was significantly increased. Moreover, the grasp tasks have been becoming more challenging, from grasping a single object to grasping an object in the cluster. Recent work used another data-collection method and human-labeled pictures to predict grasping candidates, and the results indicated that data-driven machine learning method was feasible for both single object grasp and cluttered objects grasp \cite{redmon2015real, chu2018real, lenz2015deep}. Simulation is an alternative way to collect a large-scale dataset, which has been proved to be valid and efficient \cite{mahler2017dex, mahler2016dex, mahler2019learning, 7254318}.

Although researchers have proposed various learning methods to tackle the grasping problem, the data collection remains a relatively expensive and time-consuming process. The large scale of data needed to train a deep neural network increases the difficulty of data collection. For instance, \cite{pmlr-v87-kalashnikov18a} extended the sample size of the dataset to millions using a fleet of self-supervised robots, which was hard to reproduce given the hardware cost. On the other hand, most of the data collected in existing researches was challenging to be shareable since each dataset was dependent on the specific robot grasping system, target objects, and the experimental environment. Some benchmarks have been proposed to establish shareable and reproducible data, which define the specific protocols for various robot manipulation tasks.

\begin{figure}[tbp]
    \begin{centering}
    \textsf{\includegraphics[width=1\columnwidth]{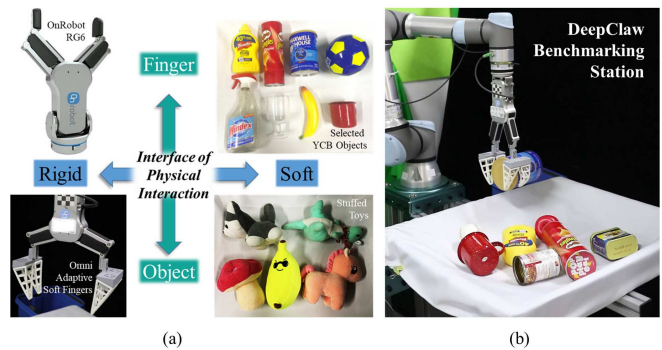}}
    \par\end{centering}
    \caption{Overview of the rigid-soft interactive learning, including (a) the four types of interaction between the fingers and objects with rigid or soft interfaces; (b) the DeepClaw benchmarking station used in this paper.}
    \label{fig:PaperOverview}
\end{figure}

In this paper, we investigate this problem from two aspects, as shown in Fig. \ref{fig:PaperOverview}. First, we propose a Rigid-Soft interactive learning method for data collection. We use soft, stuffed toys instead of daily objects for training data collection. This process can be executed without human supervision. Moreover, since the interaction between toys and fingers can reduce the complexity of motion control, a small data scale can achieve a high grasping success rate. Based on the grasp policy trained by the previous dataset, we replace the original rigid fingers with 3D-printed omni-adaptive soft fingers to deal with the Yale-CMU-Berkeley (YCB) objects, which is called Soft-Rigid interaction \cite{shintake2018soft}. The results show that such an interactive learning method can achieve a higher success grasp rate for the soft fingers with a small amount of training data on robot grasping.

Second, we aim to investigate the influence of the choice of grasp types, which means different finger configurations, on the robot learning method \cite{Cinieaau9757, nakamura2017complexities}. We explore whether the different finger configurations are transferable and whether the dataset based on different hardware is shareable. In our experiment, we use the two-parallel-finger gripper to represent a precise grasp and the four-parallel-finger gripper to represent a power grasp. Experiments show that the combination of different grasp types may contribute to a better learning policy. We find that using the precise grasp for data collection can acquire more accurate grasping information than the power grasp, and using the power grasp for testing achieves much higher success rate than the precise grasp, so the learning policy combining these two grasp types achieves the best performance. 

The contributions of this paper are as the following: 
\begin{enumerate}
    \item A 3D-Printed omni-adaptive soft finger which performs grasp with high adaptive ability.
    \item A novel learning method, namely Rigid-Soft interactive learning, utilizes a small amount of dataset, and transfer learning between soft and rigid objects.
    \item Experimental investigations of the Rigid-Soft interaction with precise grasp, power grasp, and the combination of them.
\end{enumerate}

In the rest of this paper, section \ref{sec:PreviousWork} briefly reviews the related literature of this work in learning-based robotic grasping and benchmarking. Section \ref{sec:Methodology} explains the proposed method of Rigid-Soft interactive learning in this paper. Experimental results are enclosed in section \ref{sec:Results}, which is followed by section \ref{sec:Conclusion} that ends the paper.

\section{Previous Work}
\label{sec:PreviousWork}
\subsection{Robot Manipulation Benchmark}

A widely accepted benchmark fosters reproducible research and can drive scientific progress. However, for robotic research, especially for robotic manipulation, benchmarking performance remains a significant challenge since both algorithms and hardware can make an influence on the final result. The YCB object dataset is a set of daily objects that cover a wide range of aspects of robotic manipulation tasks \cite{calli2015ycb}. Furthermore, the YCB authors proposed several well-defined tasks, such as pouring water, pick, and place. Amazon Picking Challenge \cite{correll2016analysis} and ACRV Picking Benchmark \cite{leitner2017acrv} focused on the full task, which defines the whole working environments, procedures, and the final objectives. They both measured the robotic grasping performance by the final results. ACRV provided a robot picking benchmark, which was similar to Amazon Picking Challenge with a well-defined object setup. REPLAB proposed a well-defined benchmark that consisted of task definition, evaluation protocol, and hardware setup, including robot system \cite{8794390}. REPLAB is meant to be reproducible since all hardware parts of the benchmark are low-cost and available, and the software is open source. 

\subsection{Learning for Grasp Planning }

There has been a recent paradigm shift in robotic grasp planning to data-driven learning method \cite{bohg2013data}. Similar to the learning methods used in computer vision, human-labeled images with bounding boxes indicating which position is graspable are feed as the input of the neural network. For example, Lenz et al. \cite{lenz2015deep} created Cornell grasping dataset, which contains over 1000 images with human labels. Based on this work, Chu et al. \cite{chu2018real} extended this dataset to multi-object, multi-grasp, and achieved over 96\% predict accuracy. However, human labeling is an expensive process, and human bias by semantics can influence the labeling result. In order to solve this problem, Mahler \cite{mahler2017dex, mahler2019learning} created a series of Dex-Net trained from a synthetic dataset of millions of point clouds, grasps, and analytic grasp thumbnails generated from thousands of 3D models in randomized poses on table.

Another solution is to optimize prediction policy in physical trials directly. For example, Pinto and Gupta \cite{pinto2016supersizing} created 50K data points collecting over 700 hours of robot grasping attempts based on trial and error experiments. Levine et al. \cite{levine2018learning} scaled up the dataset to 800k data points with a series of continuously running robotic arms.

\subsection{Rigid-Soft Interactive Learning}

The process of grasping can be complicated, sometimes involving collisions and adjustments. A straight-forward grasp planning may involve high-level motion control and high-resolution information from sensors. As for human, contact-guided placing is standard, and error recovery is quick when it is necessary at all \cite{nakamura2017complexities}. As for robots, it is difficult to have a precise prediction because of the limited precision of the camera, the limited information of the objects gained from sensors, and the complexities of robot control. Kalashnikov et al. \cite{pmlr-v87-kalashnikov18a} proposed an effective policy for closed-loop control, whereby the robot continuously updates its grasp strategy based on the most recent observations. As the first touch between the robot system and its surrounding real-world environment, robotic grippers play a significant role in the interaction. Moreover, soft robotic grippers have been proved to be adaptive and safe in the collision on a broad range of objects \cite{shintake2018soft}. So passive-adaptation through Rigid-Soft interaction may be another solution for contact complexities.

\subsection{The Choice of Grasp Type and Modular Gripper}

Every object can be grasped in several ways, and the final choice of grasp can make a significant influence on grasp performance. Cini et al. \cite{Cinieaau9757} defined a series of taxonomy used to classify grasps. The proposed taxonomy comprises three top-level categories: power, intermediate, and precision grasps. Humans can effortlessly manipulate objects and their environment, while robots are still far away from accomplishing these actions because of the control complexities, the limited sensor data, and motion inaccuracy. Modular finger configurations can partially represent different grasp types. For instance, Yale OpenHand Project \cite{ma2017yale} proposed various gripper designs, all of which used a similar body structure. Through different configurations of their modular fingers, they designed Model T42 gripper, Model T gripper, and Model O gripper, which were intended for different grasp types individually. Dex-Net 4.0 \cite{mahler2019learning} also considered different grasp types, selecting between suction cup and two-parallel gripper to perform robust grasp.

\section{Methodology}
\label{sec:Methodology}
\subsection{Problem Formulation}

We consider the problem of learning a robot grasping policy to plan a planar parallel-jaw grasp for a rigid object or a rigid object in a cluster. We learn a function that takes as input a color image and outputs an estimate of grasp reliability. 

\subsubsection{Definition}

Let $u=(\rho,\phi)\in R^{2} \times S^{1}$ denote a parallel-jaw grasp in 2D space specified by a center $\rho=(x,y)\in R^{2}$ relative to the camera and an grasp angle in the table plane $\phi\in\left[0,pi\right)$. Let $I=R^{H\text{\texttimes}W\text{\texttimes}3}$ be a color image with height $H$, width $W$ and 3 channels RGB taken by a camera. Let $Q_{\theta}(u,I)\in[0,1]$ be a grasp reliability evaluation decided by grasp configuration $u$ and color image $I$ with parameters $\theta$.

\subsubsection{Objective}

We use the self-supervised learning method to collect data. To train our neural network, we minimize the cross-entropy loss between predicted grasp reliability and the grasp result $R$: 

\begin{equation}
    \label{eq:theta}
	\hat{\theta}=\underset{\theta\epsilon\Theta}{argmin}\stackrel[i=1]{N}{\sum}\zeta(R,Q_{\theta}(u,I))
\end{equation}

where $\zeta$ is the cross-entropy loss function and $\theta$ defines the parameters of our neural network. Our goal is to learn a grasp predict function $Q_{\theta}*$ which provides robust grasp prediction measured by grasp success rate, grasp computation time (GCT) $t_{c}$, grasp execution time $t_{e}$ and mean picks per hour (MPPH), the number of objects that are successfully grasped per hour. MPPH is decided by success rate, $t_{c}$ and $t_{e}$. So the objective is to: 

\begin{equation}
    \label{eq:MPPH}
	max(MPPH)=max\sum\frac{Q_{\theta}(u,I)}{(t_{c}+t_{e})}
\end{equation}

\subsection{Grasp Learning Policy}

\subsubsection{Network Design}
Since CNN performs better in classification tasks rather than regression, we divide grasp angle $\phi\in[0, pi)$ into 18 angular bins, and a CNN model predicts the successful grasp probabilities independently for $\phi=0, 10, ..., 170$ degrees. Therefore, our problem can be thought of as an 18-way binary classification problem \cite{pinto2016curious}.

We build a fully convolutional neural network (FCN) converting from AlexNet with the following architecture: the first five convolutional layers are taken from AlexNet, followed by a $6\times6\times4096$ (kernel size $\times$ number of filters) convolutional layer, a $1\times1\times1024$ and a $1\times1\times36$ fully convolutional layers in sequence. The first five convolutional layers are initiated with weights pre-trained on ImageNet and are not trained with our dataset.

During training time, since each training data entry only has the label corresponding to one active binary classification among the 18 angular classes, the loss function is defined to compute cross-entropy of the active angular class. This is achieved by defining a mask from the grasp angles to filter out the non-active outputs of the last FCN layer, resulting in a FCN output of batch size$\times2$.

During the testing time, we can do sliding window sampling across the full image and predict the a grasp pose for each sampled patch. Though the FCN is trained on the cropped patch with a single grasp, it can be applied to inference the entire image of any size and give relatively dense predictions pixel-wise at one time. This strategy removes the need for sampling grasp configurations and significantly reduces the grasp computation time (GCT) \cite{satish2019policy}. The stride of the dense predictions equals to the multiplication of all the strides in the convolutional and max-pooling layers, which is 32 in our network architecture.

\subsubsection{Training Data Preparation}

We collect data using trial and error experiments, namely random grasps. The workspace is set up with multiple objects placed on a white background. First, we sample a grasp candidate $u=(\rho,\phi)\in R^{2}\times S^{1}$ randomly or using some sampling policy. Then, following the grasp configuration, the robot executes grasp and records the reward, whether the grasp is successful or not based on vision detection, which is similar to \cite{pinto2016supersizing}.

Given a grasp configuration $u=(\rho,\phi)$ and a corresponding color image, we crop a $160\times160$ patch centered at $\rho=(x,y)$, which covers the projection of our gripper fingertips to the image plane. The patch is resized to $227\times227$, which is the input size of the AlexNet. Hence each training data entry consists of a cropped image, a rotation angle $\phi$, and the grasp result of success or failure.

\subsection{Gripper Design}

\subsubsection{Omni-Adaptive Finger}

\begin{figure}[htbp]
    \begin{centering}
    \textsf{\includegraphics[width=1\columnwidth]{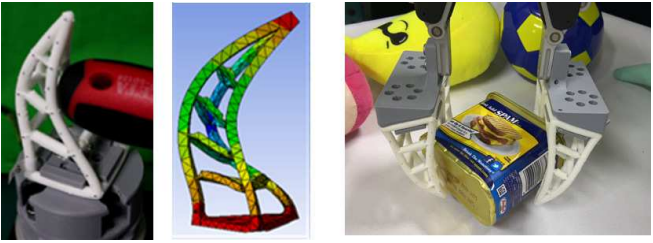}}
    \par\end{centering}
    \caption{The omni-adaptive soft finger used in this paper, including design principal, FEM simulation, and gripper integration from left to right.}
    \label{fig:OmniAdaptiveFinger}
\end{figure}

In this paper, we design a novel soft finger structure with passive Omni-Directional adaptation, ultra-low-cost, and high environmental suitability, which is shown in Fig. \ref{fig:OmniAdaptiveFinger}. This finger is inspired by Fin Ray Effect \cite{anwar2019modeling} and has excellent performance of passive adaptivity in all directions with ultra-simple structure. Additionally, this soft finger can be easily integrated with our original OnRobot RG6 gripper. 

\subsubsection{Modularity}

Our soft finger has excellent modularity in design, which can be reconfigured in a short time. Also, due to the passive-adaptivity, our finger can be easily mounted on the fingertips of existing grippers without electronic connection. So we use soft fingers in replace of the rigid RG6 fingertips when grasping YCB objects. There are two soft gripper configurations tested in this paper: two-parallel-finger configuration representing precise grasp and four-finger configuration with two fingers on each side representing power grasp. 

\subsubsection{Rigid-Soft Interactive learning}

Common interaction types between fingers and objects can be divided into four categories: Rigid-Rigid interaction, which is the case for majorities of grasping research \cite{pinto2016supersizing, pinto2016curious}, Rigid-Soft Interaction, Soft-Rigid Interaction, and Soft-Soft Interaction. Soft-Soft interaction means both the fingers and the objects are soft, and the deformations between them are complicated, which is beyond the scope of this paper.

\begin{figure}[htbp]
    \begin{centering}
    \textsf{\includegraphics[width=1\columnwidth]{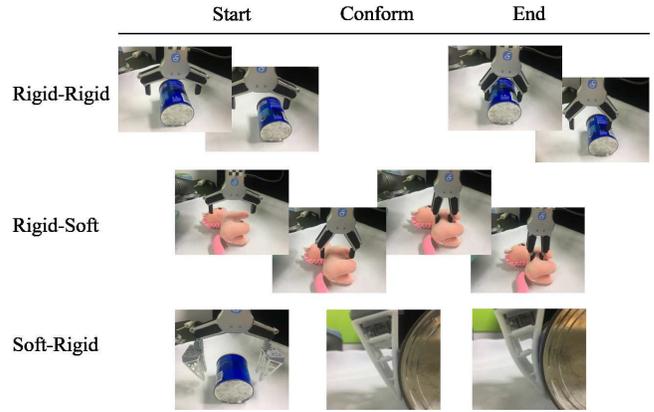}}
    \par\end{centering}
    \caption{The process of physical interactions between the rigid and soft components and between the fingers and objects during grasping, in which the Rigid-Rigid interaction may cause emergency stop. Rigid-Soft and Soft-Rigid interactions involve conformation on either objects or grippers.}
    \label{fig:ConformProcess}
\end{figure}

As for the other three categories, we divide the complete grasp into three sub-processes: start, conform, stop. As shown in Fig. \ref{fig:ConformProcess}, Rigid-Rigid interaction needs a precise prediction on the grasp point and grasp angle, especially for the objects with complex shapes. However, due to the imprecision brought by the camera, calibration, robot controller, sampling method, it is difficult to have a precise prediction. With bad grasp predictions, collisions between rigid fingers and rigid objects can result in emergency stops of the robot and irreversible damages of fingers and objects. So it is difficult to collect data using trial and error experiments based on rigid-rigid interaction. 

Rigid-Soft interaction means using rigid fingers to grasp soft objects, which is similar to the arcade claw machine. Inspired by this, we use rigid fingers to grasp soft toys with different sizes and colors, which is a relatively easy task because of the soft nature of toys. On the other hand, Soft-Rigid Interaction means using soft fingers to grasp rigid objects, which is similar to human grasping. Human hands have an excellent adaptivity over all kinds of objects with different shapes and sizes.

Overall, during the physical interaction between the rigid and soft components, the Soft-Rigid and Rigid-Soft interactions have similar performance, one of the surfaces conforms to adapt to the other surface. Also, these two interaction types have their own characteristics: soft toys are easy to grasp and can be grasped along various directions as long as the gripper pinches some parts of the soft tissues. One can collect 3000 grasps data within 10 hours, which is an efficient and safe way for data collection. Although grasping soft toys will cause the defection of predicting grasp angle, it can be used to guide our further training on grasping YCB rigid objects. A combination of these two interaction types can lead to a better performance than using the two types individually.

\section{Experiments}
\label{sec:Experiments}
\subsection{DeepClaw Benchmark}

To be reproducible, we adopted DeepClaw Benchmark to define our object set, robot grasping system, workflow, and measuring metrics. 

\subsubsection{Grasp Objects}

In this paper, we used three sets of objects, including (1) Soft toys object set: 25 soft toys with all kinds of shapes, sizes, and colors for the first stage of our training process; (2) Level-1 YCB object set consisting of 8 YCB objects, with easy-to-grasp features such as medium size and regular shape; and (3) Level-2 YCB object set consisting of 8 YCB objects with hard-to-grasp features such as relatively large size and irregular shape.

\begin{figure}[htbp]
    \begin{centering}
    \textsf{\includegraphics[width=1\columnwidth]{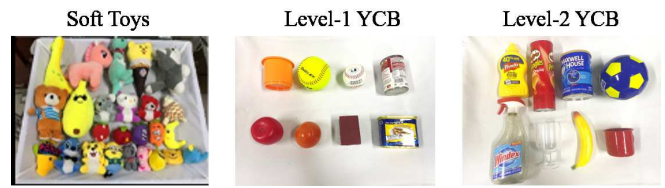}}
    \par\end{centering}
    \caption{Sets of soft toys and Level-1 YCB objects and Level-2 YCB objects.}
    \label{fig:RobotSystem}
\end{figure}

\subsubsection{Robot Grasping System }

Our experiments were carried out on a UR5 robot with an OnRobot RG6 gripper. In some experiments, soft fingers were mounted on the RG6 gripper replacing the original rigid fingertips. As shown in Fig. \ref{fig:RobotSystem}, objects were placed in the white bin, and the task was to pick the objects from the white bin and place them into the blue bin. A Realsense D435 depth camera was mounted about one meter above the workspace and provided a $1280\times720$ resolution RGB image. All experiments ran on Ubuntu 16.04 with a 2.7GHz Intel Core i7-7700HQ CPU and an NVIDIA GeForce GTX1060.

\subsubsection{Workflow of Tasks}

In this paper, we considered two types of tasks: single-task and full-task. The goal of the single-task is to grasp and transport an object to a receptacle. The goal of the full-task is to sequentially grasp all objects in the clutter and clear the workspace within limited attempts. 

\begin{figure}[htbp]
    \begin{centering}
    \textsf{\includegraphics[width=1\columnwidth]{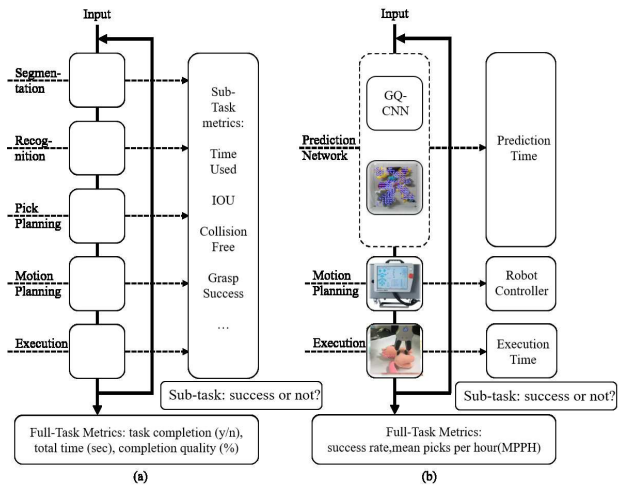}}
    \par\end{centering}
    \caption{The functional workflow of DeepClaw: (a) with standard function; (b) the one used in this project}
    \label{fig:DeepClaw}
\end{figure}

As for both single-task and full-task, there are several sub-tasks: Segmentation, Object classification, Picking Planning, Motion Planning, and Execution. In this paper, we only considered the problem of pick planning and execution measured by Grasp Success Rate, Grasp Computation Time (GCT) $t_{c}$, grasp execution time $t_{e}$ and mean picks per hour (MPPH), as shown in Fig. \ref{fig:DeepClaw}.

\subsection{Training Dataset Collection}
We collected a series dataset of 5K grasp attempts, whose configurations were explained in Table \ref{tab:DataCollection}. The objects were initially placed in the white bin randomly and were placed back randomly at the end of the grasp attempt. The procedures were automatic with minimum human interventions. In the rest of the paper, the dataset will be refer to in the format of Gripper-Object combination. For example, 4-finger Soft-Rigid guided dataset means the grasp attempts in the dataset were predicted by grasp policy trained from the Rigid-Soft dataset and were executed using a 4-soft-finger gripper to grasp the YCB objects. The Rigid-Rigid dataset was collected with a force sensor to avoid collision during the grasp attempts. Among the 500 attempts, 136 were success, 233 were failure while the rest were not completed due to collision. The Rigid-Rigid dataset was difficult to scale-up and had the risk of damaging the objects and the gripper, which is not used in the following testing experiments. The 4-finger Soft-Rigid guided and Rigid-Soft dataset had the highest success rates among the five configurations, which were 46.5\% and 38.9\% respectively. The 4-finger Soft-Rigid and 2-finger Soft-Rigid guided were about the same, which were 25.2\% and 26.2\% respectively.

\begin{table}[htbp]
    \caption{Data collection setup between rigid and soft properties of the object and fingers with different sets of fingers.}
    \label{tab:DataCollection}
    \begin{tabular}{ccc}
    \hline
    \multicolumn{1}{c|}{\textbf{\begin{tabular}[c]{@{}c@{}}Data \\ Collection\end{tabular}}}                                                                           & \multicolumn{2}{c}{\textbf{\begin{tabular}[c]{@{}c@{}}Object Property\\ Rigid: YCB, Soft: Toys\end{tabular}}}                                                             \\ \hline
    \multicolumn{1}{c|}{\multirow{4}{*}{\textbf{\begin{tabular}[c]{@{}c@{}}Finger \\ Property\\ Rigid: RG6\\ Soft: Omni\end{tabular}}}}                                & \multicolumn{1}{c|}{2Finger-RigidRigid-0.5K}                                                      & 2Finger-RigidSoft-5K                                                  \\ \cline{2-3} 
    \multicolumn{1}{c|}{}                                                                                                                                              & \multicolumn{1}{c|}{4Finger-SoftRigid-5K}                                                         & \multirow{3}{*}{None}                                                 \\
    \multicolumn{1}{c|}{}                                                                                                                                              & \multicolumn{1}{c|}{4Finger-SoftRigid-2.5K-Guided}                                                &                                                                   \\
    \multicolumn{1}{c|}{}                                                                                                                                              & \multicolumn{1}{c|}{2Finger-SoftRigid-2.5K-Guided}                                                &                                                                       \\ \hline
    \multicolumn{3}{l}{\textit{\begin{tabular}[c]{@{}l@{}}Notes:\\ 2Finger/4Finger: the number of fingers on the gripper;\\ RigidRigid/RigidSoft/SoftRigid: the properties of Finger-Object pair;\\ 2.5K/5K: the number of picking trails collected in the data;\\ Guided: whether the data collection process used guided picking.\end{tabular}}}
    \end{tabular}
\end{table}

\subsection{Testing Experiments}
To investigate the effect of the Rigid-Soft interactive learning method and soft finger configuration, we trained four grasp policies and tested their robustness with a 4-soft-finger gripper grasping YCB objects as described in tests 1 to 4 in Table \ref{tab:Experiment1}. We also explored the transfer learning method for grasp type by benchmarking grasp policies and finger configurations. We used two-parallel-fingers and four-finger configurations to represent precise grasp and power grasp. Tests 4 to 7 in Table \ref{tab:Experiment1} used a 2-soft-finger or a 4-soft-finger gripper to collect training data and a 2-soft-finger or a 4-soft-finger gripper to test, leading to 4 possible combinations. For tests 1 to 7, each of the sixteen YCB objects was placed in the bin, and the robot tried ten grasp attempts. The success rate of each test was then averaged over ten attempts of each object and further averaged over level-1 and level-2 YCB objects, respectively.

\begin{table}[htbp]
    \centering
    \caption{Experiment setup for robustness tests of grasping  policies  trained  from  different  dataset. Tests were all conducted to grasp YCB objects using soft fingers.}
    \label{tab:Experiment1}
    \begin{tabular}{c|c|c}
    \hline
    \textbf{Test \#} & \textbf{Training Dataset (5K in total)}                                                                   & \multicolumn{1}{l}{\textbf{Test Finger Setup}} \\ \hline
    \textbf{1}       & 100\%: 2Finger-RigidSoft-5K                                                                               & \multirow{5}{*}{4Finger}                       \\ \cline{1-2}
    \textbf{2}       & 100\%: 4Finger-SoftRigid-5K                                                                               &                                                \\ \cline{1-2}
    \textbf{3}       & \begin{tabular}[c]{@{}c@{}}50\%: 2Finger-RigidSoft-5K\\ 50\%: 4Finger-SoftRigid-5K\end{tabular}           &                                                \\ \cline{1-2}
    \textbf{4}       & \begin{tabular}[c]{@{}c@{}}50\%: 2Finger-RigidSoft-5K\\ 100\%: 4Finger-SoftRigid-2.5K-Guided\end{tabular} &                                                \\ \cline{1-2}
    \textbf{5}       & \begin{tabular}[c]{@{}c@{}}50\%: 2Finger-RigidSoft-5K\\ 100\%: 2Finger-SoftRigid-2.5K-Guided\end{tabular} &                                                \\ \hline
    \textbf{6}       & \begin{tabular}[c]{@{}c@{}}50\%: 2Finger-RigidSoft-5K\\ 100\%: 4Finger-SoftRigid-2.5K-Guided\end{tabular} & \multirow{2}{*}{2Finger}                       \\ \cline{1-2}
    \textbf{7}       & \begin{tabular}[c]{@{}c@{}}50\%: 2Finger-RigidSoft-5K\\ 100\%: 2Finger-SoftRigid-2.5K-Guided\end{tabular} &                                                \\ \hline
    \end{tabular}
\end{table}

\section{Results}
\label{sec:Results}
\subsection{Rigid-Soft Interactive learning}
\begin{figure}[htbp]
    \begin{centering}
    \textsf{\includegraphics[width=1\columnwidth]{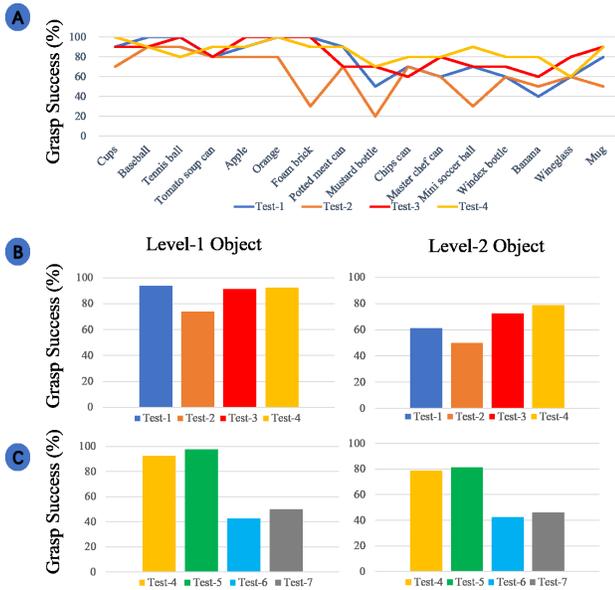}}
    \par\end{centering}
    \caption{Performance of different learning policy and finger configurations. Success rate of (A) each selected YCB
    objects in test 3 averaged over 10 grasp attempts, (B) Level-1 and Level-2 Object in test 1 to 4, (B) Level-1 and Level-2 Object in test 4 to 7.}
    \label{fig:LearnedResults}
\end{figure}

Fig. \ref{fig:LearnedResults}(A) shows the grasp success rate for each YCB objects in tests 1 to 4. The success rates dropped about 20\% for level-2 objects in comparison to level-1 objects except test 4 with guided training dataset. In general, objects with round-like shape and medium size were easier to grasp, for example all the ball except mini soccer ball, plastic apple and even mug. Soft ball and baseball are the two heaviest objects among the sixteen YCB objects, yet they had comparable success rates as much lighter plastic apple and orange. The more difficult objects like banana and mustard bottle have the lowest height when laid in the bin, hence the adaptation of the soft finger is very limited.

As shown in Fig. \ref{fig:LearnedResults}(B), test 3 with combination of Rigid-Soft and 4-finger Soft-Rigid training dataset improved the grasping performance in comparison with pure Rigid-Soft or 4-finger Soft-Rigid dataset alone, achieving 91.25\% success rate for Level-1 objects and 72.50\% success rate for Level-2 objects. Further more, in test 4, 4-finger Soft-Rigid guided data helped to improve the success rate for level-2 objects to almost 80\% while not helping for level-1 objects. This result suggested guided grasp improves the quality of the training dataset for difficult objects.

\subsection{Effects of Grasp Types}
Fig. \ref{fig:LearnedResults}(C) displays the results for tests 4 to 7 which were also named power, power-precise, precise-power and precise grasps according to their soft finger configurations during training and testing phases. In general, 4-finger gripper performed much better than 2-finger gripper, in spite of different grasp policies trained from different dataset. The success rates of 4-finger gripper were twice the success rates of 2-finger gripper. Another interesting result was that the policy trained from 2-finger Soft-Rigid guided dataset performed better than that trained from 4-finger Soft-Rigid guided dataset, in spite of being executed with the 2-soft-finger gripper or the 4-soft-finger gripper. In All, Precise-Power group achieved the highest success rate at 97.5\% for level-1 objects and 81.3\% for level-2 objects, which used the two-soft-finger gripper to collect train data and the four-soft-finger gripper to test.

\subsection{Effect of data size}

\begin{figure}[htbp]
    \begin{centering}
    \textsf{\includegraphics[width=1\columnwidth]{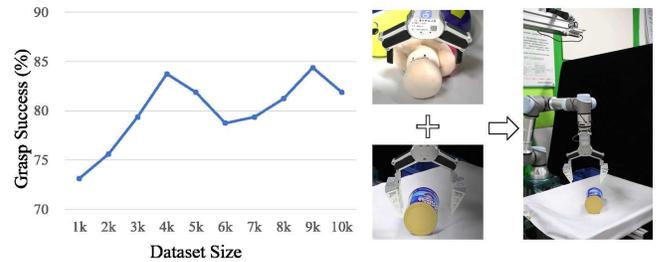}}
    \par\end{centering}
    \caption{Comparison of the performance of policies trained from different training set sizes. The improvement stops after 4k attempts in our grasping scenario.}
    \label{fig:DatasetSize}
\end{figure}

We extended the test 3 to investigate the effect of training data size, which all consisted of equal grasp attempts from Rigid-Soft and 4-finger Soft-Rigid dataset. As shown in Fig. \ref{fig:DatasetSize}, success rate initially increased as data size increased and saturated quickly under 85\% after 4000 attempts. More data did not improve the reliability further. The saturation success rate is partly limited by the gripper itself and can be further promoted with improved soft gripper designs and similar size of training data. In the current design of the soft finger, the finger configurations and frictions between the gripper and the object are not in the main scope of this paper, and will be fully explored in a following paper by the authors.

\subsection{Comparison with heuristic baseline}

We compared our Rigid-Soft interactive learning method with a simple heuristic baseline. The heuristic grasping rules locate the minimum bounding box of the object from its contours and grasp about the center of the bounding box along direction of the shorter edge of the bounding box. In order to differentiate the performance improvement by the soft fingers and the learning methods, we conducted experiments using the heuristic baseline with RG6 gripper and the soft fingers respectively. RG6 achieved 53.8\% and 55\% while 4-soft-finger gripper achieved 78.8\% and 55\% for level-1 and level-2 objects respectively. The soft fingers improved the performance on the level-1 objects thanks to its embracing ability for round-like shape and medium objects but did not improve the performance on the level-2 objects on average. The proposed learning method further promoted the success rate to 97.5\% and 81.3\% for level-1 and level-2 objects. The boost is mainly contributed by the Rigid-Soft interactive learning method.

\subsection{Clutter Removal}

\begin{figure}[htbp]
    \begin{centering}
    \textsf{\includegraphics[width=1.0\columnwidth]{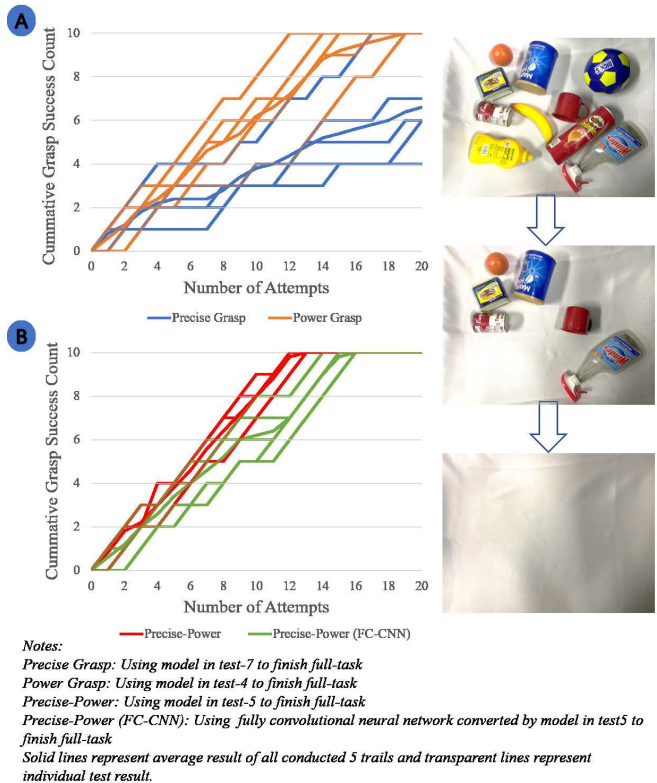}}
    \par\end{centering}
    \caption{Experiment results for the cluster removal full task of 5 trials of 20 attempts each trail.}
    \label{fig:SuccessRate}
\end{figure}

Since our training data collection involved objects in clutter, the trained policies should also be able to complete our full-task: clutter removal. We tested four experimental setups to remove a clutter of ten objects, five objects from Level-1 object set and five objects from Level-2 object set. As shown in Fig. \ref{fig:SuccessRate}, Power, Precise-Power and Precise used the same setup as tests 4, 5, and 7 respectively. Precise-Power group achieved the highest success rate. Table \ref{table:ResultSummary} shows the detailed performance on cluster removal task. Although the fully-convolutional neural network does not perform as well as the sampling-based network, it significantly reduced computation time hence achieved the highest MPPH. 

\begin{table}[htbp]
    \caption{Overall experiment results when using different grasp types and learning methods in the cluster removal task.}
    \label{table:ResultSummary}
    \centering
    \resizebox{\columnwidth}{!}{%
    \begin{tabular}{ccccccc}
    \hline
    \textbf{Setup}                                                                & \textbf{\begin{tabular}[c]{@{}c@{}}$t_c$ \\ (s)\end{tabular}} & \textbf{\begin{tabular}[c]{@{}c@{}}$t_e$ \\ (s)\end{tabular}} & \textbf{\begin{tabular}[c]{@{}c@{}}Success \\ Rate (\%)\end{tabular}} & \textbf{MPPH}   & \textbf{Attempts} & \textbf{Failures} \\ \hline
    \textbf{\begin{tabular}[c]{@{}c@{}}Precise \\ Grasp\end{tabular}}             & 10.3                                                          & 11.2                                                          & 35.05                                                                 & 104.34          & 97                & 34                \\ \hline
    \textbf{\begin{tabular}[c]{@{}c@{}}Power \\ Grasp\end{tabular}}               & 10.1                                                          & 10.4                                                          & 66.67                                                                 & 208.13          & 75                & 25                \\ \hline
    \textbf{\begin{tabular}[c]{@{}c@{}}Precise-\\ Power\end{tabular}}             & 10.2                                                          & 10.3                                                          & \textbf{81.97}                                                        & 255.90          & \textbf{61}       & \textbf{11}       \\ \hline
    \textbf{\begin{tabular}[c]{@{}c@{}}Precise-\\ Power \\ (FC-CNN)\end{tabular}} & \textbf{0.16}                                                 & 10.4                                                          & 74.63                                                                 & \textbf{452.28} & 68                & 18               
    \end{tabular}%
    }
\end{table}

\section{Discussions}
\label{sec:Discussions}
\subsection{Rigid-Soft Interactive Learning}

The experiment results shown in Fig. \ref{fig:LearnedResults}(B) highlight the advantage of Rigid-Soft interactive learning. Although the Rigid-Soft dataset with soft toys achieved high reliability on Level-1 objects, it only achieved about 60\% reliability on Level-2 objects, which consists of some rectangular items. The plausible reason is that the soft surface of toys has a high tolerance on the grasping directions, which means this dataset retains useful location information but loses most of the orientation information of the objects. The observation also confirms this hypothesis that the grasp orientations predicted by the grasp policy trained from the Rigid-Soft dataset are rather uniform among different objects and locations. 

On the other hand, the performance of the 4-finger Soft-Rigid dataset with YCB objects in test 2 was unexpectedly bad, probably because it cannot precisely localize the interest area. We have observed from the experiments that the grasp location predicted by the grasp policy trained from the 4-finger Soft-Rigid dataset was not very stable and sometimes not optimal. By combining the two datasets in test 3, we have achieved a higher success rate than both datasets alone with the same amount of data. This result suggests that the Soft-Rigid dataset retains the orientation information better than the Rigid-Soft dataset while the other way around for location information. In general, it might be a good practice to combine training data from different grippers and objects to leverage the advantage of each data source.

\subsection{Modular Gripper System for Reconfiguration}

The results in Fig. \ref{fig:LearnedResults}(C) and Fig. \ref{fig:SuccessRate} also suggest that finger configuration has an essential effect on the final performance. The 4-soft-finger gripper has a stronger loading performance, especially when grasping spherical or large items and outperforms 2-soft-finger gripper significantly. Also, this configuration has a better fault tolerance, which means a little fault in the prediction pose may not influence grasp performance. However, high fault tolerance leads to an ambiguous effect during the data collection process, and in return, lowers the performance of the trained policy, as shown in test 2. In contrast, the 2-finger Soft-Rigid dataset seems to retain more accurate grasp information than 4-finger. Hence, using the 2-soft-finger gripper to collect data and the 4-soft-finger gripper to test appears to be the best combination.  

\subsection{Fully Convolutional Network and Edge Computing}

As shown in Table \ref{table:ResultSummary}, with the same density of predicted grasps, the fully convolutional network can significantly reduce the computation time since it can eliminate the sampling process and predict a dense map of grasp poses at one time. Although the success rate of FCN is lower slightly, the reduced computation time helps it achieve the highest MPPH. 

\section{Conclusion}
\label{sec:Conclusion}
We have presented a novel soft finger design capable of omni-directional adaptation when grasping the objects with an uneven surface. The proposed Rigid-Soft interactive learning method achieves good results with a small amount of training data. We can train a network predicting grasp position and angle within several days. We compare grasp policies trained from the Rigid-Soft dataset, Soft-Rigid, and their combinations. Besides, the results show that each dataset has its advantage and disadvantage, and a combination of them achieves the best reliability in terms of success rate. It is also found that different types of grasping influence training performance. By changing the finger configuration from two soft fingers to four soft fingers, the grasping success rate can be improved significantly.

This work is a preliminary exploration of the grasp types of soft fingers. In future research, we would like to further investigate the influence of soft finger configuration and design on robot learning. In particular, we would like to consider designing a complete modular gripper system with adaptive fingers, a machine vision system, and a computation system. Edge computing is another direction to distribute the learning capabilities for localized decision making in robotic grasping. We intend to integrate an embedded AI computing platform, like Jetson TX2, for this purpose in future research.

                                  
\section*{ACKNOWLEDGMENT}
We would like to express our thanks to Dr. Aaron Dollar, Dr. Cali Berk, and Dr. Adam Norton for providing the YCB objects for testing, as well as the valuable comments received from the ICRA 2019 workshop on Benchmarking Robotic Manipulations. We would also like to acknowledge the funding support by the SUSTech Faculty of Engineering (No.: 34/K18341702) and the AncoraSpring, Inc., Shenzhen.
\bibliographystyle{IEEEtran}
\bibliography{ref-InteractiveLearn}



\end{document}